\documentclass[journal, twoside]{IEEEtran}

\pdfoutput=1 

\usepackage{times}
\usepackage{epsfig}
\usepackage{graphicx}
\usepackage{amsmath, amsthm}
\usepackage{amssymb}
\usepackage{float, array}
\usepackage{multirow}
\usepackage{epstopdf}
\usepackage{subfigure, color, cite}

\begin{document}
%
\title{SADet: Learning An Efficient and Accurate Pedestrian Detector}
%
%
\author{Chubin Zhuang, Zhen Lei, Stan Z. Li
\thanks{C. Zhuang, Z. Lei and S.Z. Li are with the Center for Biometrics and Security Research and National Laboratory of Pattern Recognition, Institute of Automation, Chinese Academy of Sciences, Beijing 100190, China, and with the University of Chinese Academy of Sciences, Beijing 100049, China. E-mail: \{zlei, chubin.zhuang, szli\}@nlpr.ia.ac.cn.}
}

\maketitle

\begin{abstract}
  Although the anchor-based detectors have taken a big step forward in pedestrian detection, the overall performance of algorithm still needs further improvement for practical applications, \emph{e.g.}, a good trade-off between the accuracy and efficiency. To this end, this paper proposes a series of systematic optimization strategies for the detection pipeline of one-stage detector, forming a single shot anchor-based detector (SADet) for efficient and accurate pedestrian detection, which includes three main improvements. Firstly, we optimize the sample generation process by assigning soft tags to the outlier samples to generate semi-positive samples with continuous tag value between $0$ and $1$, which not only produces more valid samples, but also strengthens the robustness of the model. Secondly, a novel Center-$IoU$ loss is applied as a new regression loss for bounding box regression, which not only retains the good characteristics of IoU loss, but also solves some defects of it. Thirdly, we also design Cosine-NMS for the postprocess of predicted bounding boxes, and further propose adaptive anchor matching to enable the model to adaptively match the anchor boxes to full or visible bounding boxes according to the degree of occlusion, making the NMS and anchor matching algorithms more suitable for occluded pedestrian detection. Though structurally simple, it presents state-of-the-art result and real-time speed of $20$ FPS for VGA-resolution images ($640 \times 480$) on challenging pedestrian detection benchmarks, i.e., CityPersons, Caltech, and human detection benchmark CrowdHuman, leading to a new attractive pedestrian detector.

\end{abstract}


\ifCLASSOPTIONpeerreview
\begin{center} \bfseries EDICS Category: 3-BBND \end{center}
\fi
%
\IEEEpeerreviewmaketitle

\section{Introduction}
\IEEEPARstart{P}{edestrian} detection is a fundamental and essential step for many pedestrian related applications, \emph{e.g.}, person re-identification\cite{li2016discriminative, wu2019unsupervised, li2018harmonious}, human gait recognition\cite{liu2018learning, arshad2019multi}, and is of great requirement to both accuracy and efficiency. Following the milestone work of Viola-Jones\cite{viola2004robust}, most of the early works focus on designing robust features\cite{lowe2004distinctive, zhu2006fast} and training effective classifiers\cite{benenson2013seeking, yan2012multi}. But these approaches rely heavily on the non-robust hand-crafted features and optimize each component separately, making pedestrian detection pipeline sub-optimal.

Recent years have witnessed the remarkable success achieved by convolutional neural network (CNN)\cite{krizhevsky2012imagenet}, ranging from image classification\cite{he2016deep, simonyan2014very, szegedy2016rethinking, howard2017mobilenets} to object detection\cite{ren2015faster, liu2016ssd, zhuang2019fldet, girshick2015fast}, which also inspires pedestrian detection. R-CNN\cite{girshick2014rich} firstly applies CNN to object detection based on proposals generated by Selective Search\cite{uijlings2013selective}. Following R-CNN, Region Proposal Network (RPN) integrated with pre-defined anchors are designed to generate candidate proposals in a unified framework, forming the two-stage detector Faster R-CNN\cite{ren2015faster}, which is the originator of anchor-based methods. Beyond the success achieved by Faster R-CNN on generic object detection, numerous extensions to this framework have been proposed and demonstrate superior performance on pedestrian detection, \emph{e.g.}, Adapted Faster R-CNN\cite{zhang2016faster}. Nevertheless, these two-stage anchor-based detectors are still far from practical applications for the low run-time efficiency caused by the time-consuming two-stage processing, including proposal generation and classification of RoI pooling features.

Alternatively, aiming at higher run-time efficiency, the one-stage method (\emph{e.g.}, SSD\cite{liu2016ssd}) discards the second stage of Faster R-CNN and directly detects objects by regular and dense sampling over locations, scales and aspect ratios. Though faster, one-stage detector has not presented competitive results on common pedestrian detection benchmarks (\emph{e.g.}, CityPersons\cite{zhang2017citypersons} and Caltech\cite{dollar2011pedestrian}). Afterwards, although some improved versions (\emph{e.g.}, ALFNet\cite{liu2018learning}) based on one-stage framework with multi-step regression are proposed to find a better balance between accuracy and efficiency, the final performance of these detectors are still not that satisfactory, which gets even worse when compared with the newly anchor-free detectors (\emph{e.g.}, CSP\cite{liu2019high}).

\begin{figure*}
\begin{center}
\includegraphics[scale=0.74]{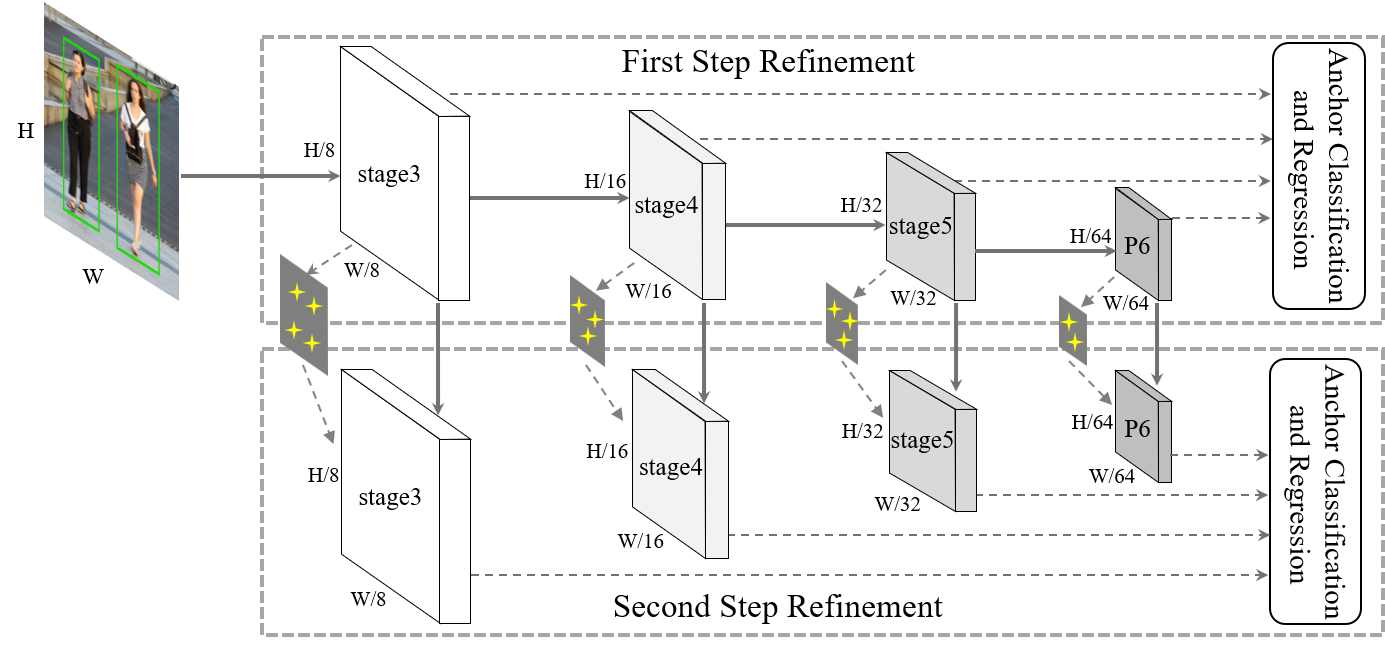}
\end{center}
\caption{Architecture of SADet. For better visualization, we only display the layers used for detection. The celadon parallelograms denote the refined anchors associated with different feature layers after the first-step regression. The stars represent the centers of the refined anchor boxes, which are not regularly paved on the image.}
\label{figure1}
\end{figure*}

In pursuit of a better balance between accuracy and efficiency for the anchor-based detector, a series of systematic optimizations are adopted for the detection pipeline of one-stage detector, including sample generation, bounding box regression loss function, Non-Maximum Suppression (NMS) algorithm and anchor matching algorithm, forming a systematic and effective framework for pedestrian detection, which can be easily extended to other anchor-based pedestrian detector. The concerns may generally come as follows: \textbf{1)} In common paradigm of sample generation process, two threshold values \{$T_{neg}$, $T_{pos}$\} are usually pre-defined to determine whether the anchor boxes are positive or negative, the anchor boxes with jaccard overlap between $T_{neg}$ and $T_{pos}$ will be discarded subsequently, which makes the model sensitive to the thresholds and cannot fully utilize all valid samples. \textbf{2)} The distance losses such as $l_1$ and $l_2$-norms are generally applied as the training objective for bounding box regression, however, as mentioned in \cite{rezatofighi2019generalized}, there exists a gap between optimizing the commonly used distance losses for regressing the parameters of a bounding box and maximizing the evaluation metric ($IoU$), which makes pedestrian localization not accurate enough. \textbf{3)} In the existing pedestrian detection pipeline, both NMS algorithm and anchor matching algorithm are directly adopted from general object detector, which is not applicable to the occlusion problem in pedestrian detection.

To handle these challenging issues, this paper proposes a series of improvements to the detection pipeline of pedestrian detector, which can be summarized as follows: \textbf{1)} We optimize the sample generation process by assigning soft tags ranging from $0$ to $1$ to these outlier samples according to their jaccard overlap, and further add these semi-positive samples to the training objective as well. In this way, not only more valid samples are generated, but also the robustness of detector is strengthened. \textbf{2)} A novel Center-$IoU$ loss is specially designed as a new regression loss for more precise pedestrian localization, which not only retains the good characteristics of $IoU$ loss, but also solves some defects of it. \textbf{3)} We design Cosine-NMS for the postprocess of predicted bounding boxes to reduce false detections of the adjacent overlapping pedestrians, and further propose adaptive anchor matching algorithm to enable the model to adaptively match the anchor boxes to full or visible bounding boxes according to the degree of occlusion, making the Non-Maximum Suppression (NMS) and anchor matching algorithms more suitable for occluded pedestrian detection. To better verify the effectiveness of the above methods, the relevant experiments are directly conducted on a very strong baseline ALFNet\cite{liu2018learning}, which owns great performance on pedestrian detection. Notably, the aforementioned optimization strategies are systematic improvements to the detection pipeline of one-stage pedestrian detector, which can also be applied to any anchor-based pedestrian detector, including one-stage and two-stage methods.

In conclusion, this paper proposes a single shot anchor-based pedestrian detector (SADet) to achieve efficient and accurate pedestrian detection. Though structurally simple, the proposed SADet presents state-of-the-art result and real-time speed of $20$ FPS for VGA-resolution images ($640 \times 480$) on challenging pedestrian detection benchmarks, \emph{i.e.}, CityPersons\cite{zhang2017citypersons}, Caltech\cite{dollar2011pedestrian}, and human detection benchmark CrowdHuman\cite{shao2018crowdhuman}, which demonstrates the supreme generalization capacity and effectiveness of our method.

For clarity, the main contributions of this work can be summarized as follows:
\begin{itemize}
\setlength{\itemsep}{1pt}
\setlength{\parsep}{1pt}
\setlength{\parskip}{1pt}
\item We introduce a new design of soft label to sample generation process to fully utilize all valid samples and improve the robustness of model.
\item We propose a novel Center-$IoU$ loss as the training objective for bounding box regression, which not only retains the good characteristics of $IoU$ loss, but also solves some defects of it.
\item We design Cosine-NMS for the postprocess of predicted bounding boxes to reduce false detections of the adjacent overlapping pedestrians, by assigning a higher penalty to the highly overlapped detections.
\item We propose adaptive anchor matching to enable the model to adaptively match the anchor boxes to full or visible bounding boxes according to the degree of occlusion, making the anchor matching algorithm more suitable for occluded pedestrian detection.
\end{itemize}

\section{Related work}
\subsection{Traditional detector}
As an important research topic in computer vision field, pedestrian detection has obtained considerable interests with extensive applications in decades. The traditional solution to this problem is training a discriminative pedestrian detector that exhaustively operates on the sub-images across all locations and scales. Dalal and Triggs\cite{dalal2005histograms} apply the histograms of oriented gradient (HOG) descriptors along with SVM (Support Vector Machine) classifier\cite{suykens1999least} to human detection. Paisitkriangkrai et al.\cite{paisitkriangkrai2014strengthening} propose a simple yet effective feature extraction method based on spatially pooled low-level visual features, and directly optimize the partial area under the ROC curve for better performance. Yan el al.\cite{yan2014fastest} introduce deformable part model (DPM)\cite{felzenszwalb2008discriminatively} into pedestrian detection task and further propose several techniques to solve the speed bottleneck of DPM, while maintaining the accuracy. ACF\cite{dollar2014fast}, LDCF\cite{nam2014local} extend the paradigm of Viola and Jones\cite{viola2004robust} to exploit various filters on Integral Channel Features (ICF)\cite{dollar2009integral} with the sliding window strategy. Zhang et al.\cite{zhang2015filtered} systematically explore different filter banks and show that they provide means for important improvements for pedestrian detection.

\subsection{Anchor-based detector}
Afterwards, coupled with the prevalence of deep learning techniques, the anchor-based methods originated from Faster R-CNN\cite{ren2015faster} rapidly dominate this field. One key component of anchor-based detectors is the pre-defined anchor boxes with fixed scales and aspect ratios around a low-resolution image grid, which will be further refined in a unified framework to get the final predictions. As the pioneering work of anchor-based methods, Faster R-CNN\cite{ren2015faster} generates candidate proposals and further classifies and refines these proposals in a two-step regression framework. In contrast, one-stage detectors, popularized by SSD\cite{liu2016ssd}, remove the proposal generation step and achieve a great balance between accuracy and efficiency. In terms of pedestrian detection, the anchor-based detectors dominate. RPN+BF\cite{zhang2016faster} adapts the original RPN in Faster-RCNN to generate proposals, then learns boosted forest classifiers on top of these proposals. MS-CNN\cite{cai2016unified} also applies the two-stage framework but generates proposals on multi-scale feature maps. Zhang et al.\cite{zhang2017citypersons} present an effective pipeline for pedestrian detection via RPN followed by boosted forests. RepLoss\cite{wang2018repulsion} and OR-CNN\cite{zhang2018occlusion} focus on the design of regression losses to tackle the occluded pedestrian detection in crowded scenes. ALFNet\cite{liu2018learning} stacks a series of predictors to directly evolve the default anchor boxes of SSD step by step into improving the detection results. Adaptive NMS\cite{liu2019adaptive} applies a dynamic suppression strategy to optimize the postprocess of predictions. Pang et al.\cite{pang2019mask} propose a Mask-Guided Attention Network (MGAN) for occluded pedestrian detection, which produces a pixel-wise attention map using visible-region information, thereby highlighting the visible body region while suppressing the occluded part of the pedestrian.

\subsection{Anchor-free detector}
Considering the tedious design of anchors, the anchor-free detectors bypass this stage and directly make predictions on an image. DenseBox\cite{huang2015densebox} first proposes a unified end-to-end fully convolutional framework that directly predicts bounding boxes. UnitBox\cite{yu2016unitbox} proposes an Intersection over Union ($IoU$) loss function for better box regression. YOLO\cite{redmon2016you} appends fully-connected layers to parse the final feature maps of a network into class confidence scores and box coordinates. Recently, with the successful application of FPN\cite{lin2017feature}, the anchor-free methods have ushered in a period of rapid development. CornerNet\cite{law2018cornernet} proposes to detect an object bounding box as a pair of corners, leading to a high performance single-shot detector. TLL\cite{song2018small} proposes to detect an object by predicting the top and bottom vertexes, which achieves significant improvement on Caltech\cite{dollar2011pedestrian}. CenterNet\cite{duan2019centernet} models an object as a single point, applies keypoint estimation to find center points and regresses to all other object properties, such as size, 3D location, orientation, and even pose. Similarly, CSP\cite{liu2019high} simplifies pedestrian detection as a straightforward center and scale prediction task through convolutions and presents great performance on pedestrian detection benchmarks CityPersons\cite{zhang2017citypersons} and Caltech\cite{dollar2011pedestrian}.

\section{Approach}
This section introduces the details of the SADet that enable the detector to be accurate and efficient on pedestrian detection, including detection pipeline, anchor design, soft label design, Center-$IoU$ loss, Cosine-NMS and adaptive anchor matching, as well as some other implementation details for training and inference.

\subsection{Detection pipeline}
In anchor-based detectors, multiple feature maps with different resolutions are extracted from a backbone network, which can be defined as follows:
\begin{equation}\label{equa1}
    \Phi_n = f_n(\Phi_{n-1})=f_n(f_{n-1}(...f_1(I))),
\end{equation}
where $I$ is the input image, $f_n(.)$ represents the $n$ th layer and $\Phi_n$ is the generated feature map. On top of these multi-scale feature maps, detection can be formulated as:

\begin{equation}\label{equa2}
    p_n(\Phi_n, B_n) = \{cls_n(\Phi_n, B_n), reg_n(\Phi_n, B_n)\},
\end{equation}
where $B_n$ is the anchor box pre-defined in the detection layer, $p_n(.)$ is the detection result of $n$ th feature map, which consists of two elements, the classification scores $cls_n(\Phi_n, B_n)$ and the related parameters for anchor box regression. Following \cite{liu2018learning}, we stack a series of predictors $p^t_n(.)$ on these anchor boxes to construct a two-step refinement framework for anchor boxes to improve the performance of one-stage detectors, which is depicted in Figure \ref{figure1}. Therefore, the above equation can be re-formulated as:

\begin{equation}\label{equa3}
    p_n(\Phi_n, B^0_n) = p^2_n(p^1_n(\Phi_n, B^0_n)),
\end{equation}

This detection pipeline contains a two-step refinement of anchor boxes, the first step is used to coarsely adjust the locations and sizes of anchors to provide better initialization for the subsequent regressor, then the second step takes the refined anchors as the input from the former to further improve the regression and classification process, and outputs the final predictions.

\begin{figure}[!ht]
\centering
\includegraphics[scale=0.5]{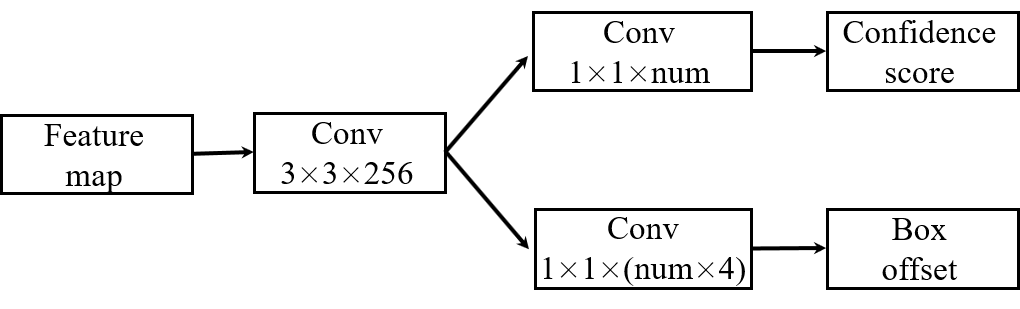}
\setlength{\belowcaptionskip}{0pt}
\setlength{\abovecaptionskip}{0pt}
\caption{The detailed implementation of the detection module applied in our SADet, which is attached to each level of feature maps to translate default anchor boxes to corresponding detection results.}
\label{figure-head}
\end{figure}

For the detection module, a set of convolutional layers are deployed to extract features from detection layers for pedestrian/non-pedestrian classification and bounding box regression, as depicted in Figure \ref{figure-head}.

\subsection{Anchor design}
Anchor-based object detectors with reasonable design of anchors in different feature maps have proven to be effective to handle objects with different scales\cite{zhang2017s3fd}. Our SADet applies ALFNet\cite{liu2018learning} as a baseline and uses ResNet-$50$\cite{he2016deep} as the backbone network of the detector, which is pictorially illustrated in Figure\ref{figure1}.

Table \ref{table1} presents the detailed anchor design of our SADet. The last layers of \textbf{stage3}, \textbf{stage4}, \textbf{stage5} in ResNet-$50$ and an additional convolutional layer \textbf{P6} attached at the end are selected as the detection layers with sizes downsampled by $8$, $16$, $32$, $64$ respectively, which are then associated with anchor boxes with width of {($16$, $24$), ($32$, $48$), ($64$, $96$), ($128$, $160$)} pixels and a single aspect ratio of $0.41$. Notably, the above settings of anchor boxes will be generally applied to all datasets without further adjustments in our experiments, including CityPersons\cite{zhang2017citypersons}, Caltech\cite{dollar2011pedestrian} and CrowdHuman\cite{shao2018crowdhuman} datasets.

\begin{table}[ht]
  \centering
  \caption{The stride size, width and aspect ratio of pre-defined anchors of the four detection layers.}\label{table1}
  \begin{tabular}{p{2.4cm}<{\centering}|p{1.0cm}<{\centering} p{1.25cm}<{\centering} p{2.0cm}<{\centering}}
    \hline
    \textbf{Detection Layer} & \textbf{Stride} & \textbf{Width} & \textbf{Aspect Ratio} \\
    \hline
    stage3 & 8  &  16, 24    & 0.41 \\
    stage4 & 16 &  32, 48    & 0.41 \\
    stage5 & 32 &  64, 96    & 0.41 \\
    P6     & 64 &  128, 160  & 0.41 \\
    \hline
  \end{tabular}
  \vspace{0.5mm}
\end{table}

\subsection{Soft label design}\label{sec-soft}
\begin{figure*}
  \centering
  \subfigure[$\{T_{neg}, T_{pos}\}$=\{$0.3$, $0.4$\}]{
  \label{soft-1}
  \begin{minipage}{5.6cm}
  \centering
    \includegraphics[width=6.5cm]{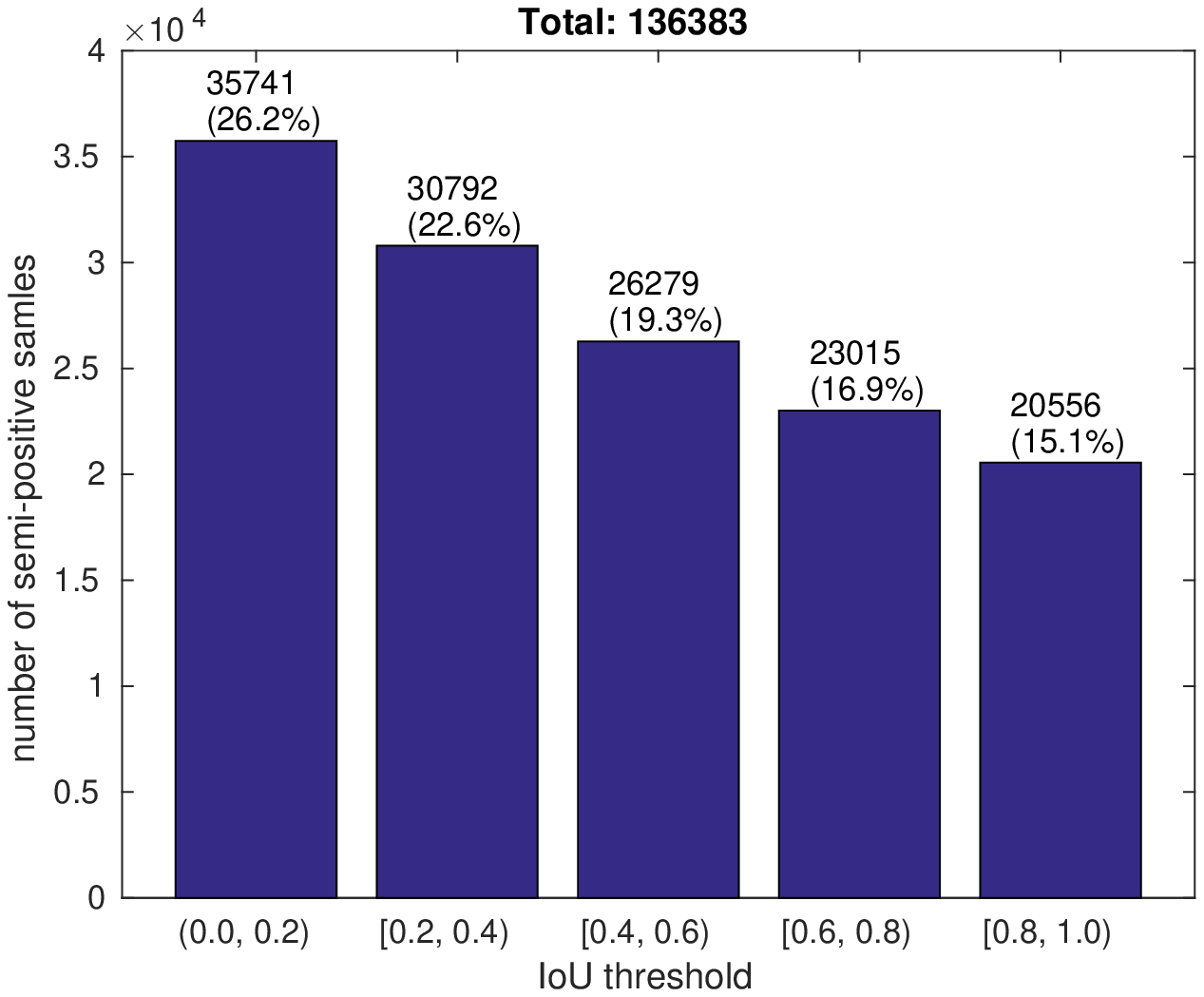}
  \end{minipage}
  }
  \subfigure[$\{T_{neg}, T_{pos}\}$=\{$0.4$, $0.5$\}]{
  \label{soft-2}
  \begin{minipage}{5.8cm}
  \centering
    \includegraphics[width=6.5cm]{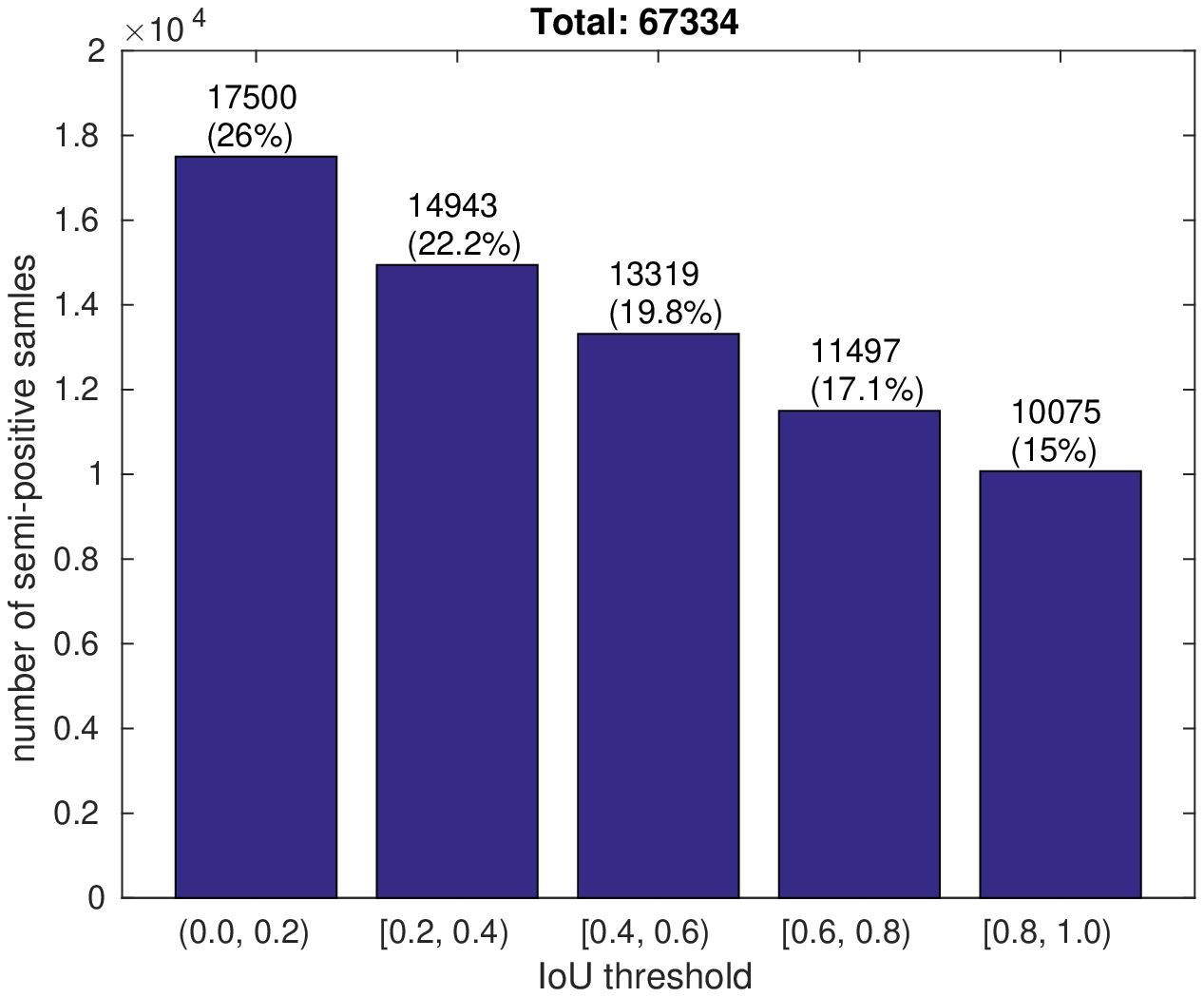}
  \end{minipage}
  }
  \subfigure[$\{T_{neg}, T_{pos}\}$=\{$0.5$, $0.6$\}]{
  \label{soft-3}
  \begin{minipage}{5.6cm}
  \centering
    \includegraphics[width=6.5cm]{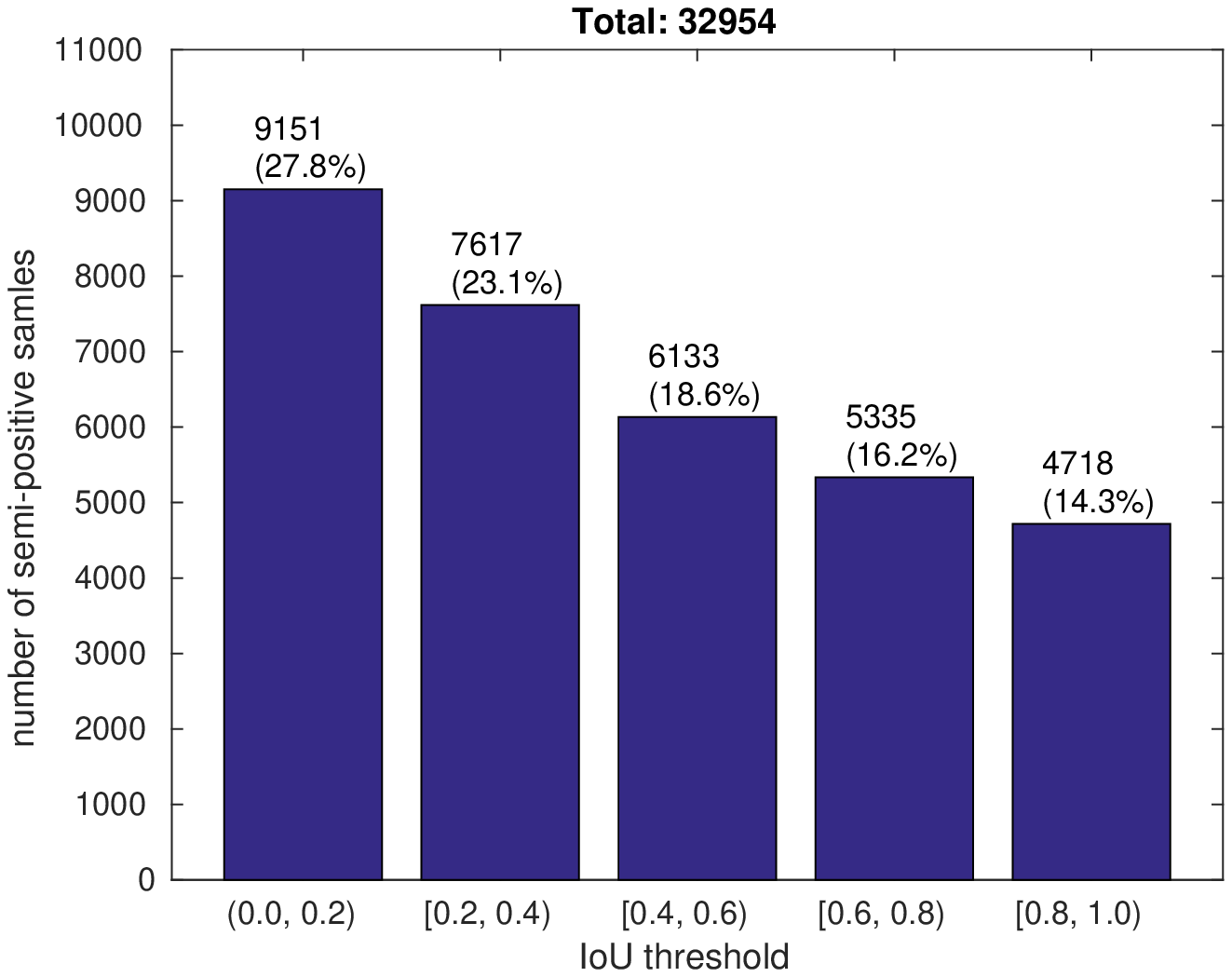}
  \end{minipage}
  }
  \caption{These bar charts depict the distribution of labels of the semi-positive samples under different $IoU$ threshold settings, \emph{i.e.}, \{$0.3$, $0.4$\}, \{$0.4$, $0.5$\} and \{$0.5$, $0.6$\} for $\{T_{neg}, T_{pos}\}$ respectively. The vertical axis represents the number of semi-positive samples generated, and the horizontal axis represents the label value of these semi-positive samples.} \label{Figure-soft}
\end{figure*}

In common paradigm of sample generation process of anchor-based methods, two threshold values \{$T_{neg}$, $T_{pos}$\} are usually pre-defined to determine whether the samples should be included in the training phase. During anchor matching phase, the anchors with $IoU$ (Intersection-over-Union) ratio higher than $T_{pos}$ with any ground-truth boxes will be assigned a positive label $1$ as the positive samples, and a negative label $0$ will be sent to a non-positive anchor if its $IoU$ ratio is lower than $T_{neg}$ for all ground-truth boxes, which is defined as a negative sample. Meanwhile, the anchors that are neither positive nor negative (with $IoU$ ratio between $T_{neg}$ and $T_{pos}$) will be ignored subsequently and do not contribute to the training objective, which makes the model sensitive to the pre-defined thresholds and cannot fully utilize all valid samples. Furthermore, the design of binary class label \{$0$, $1$\} is also not conducive to the model training, making the predictions of the model not robust enough.

Therefore, we propose a new design of soft label to utilize these ignored anchors by assigning labels ranging from $0$ to $1$ to these anchors according to their $IoU$ ratios with ground-truth boxes, and further add these samples to the training objective as well. The assignment of soft label is defined as follows.
\begin{equation}\label{equa4}
    label=
    \begin{cases}
        \quad \ 0, & \text{if\quad} iou < T_{neg} \\
        \quad \ 1, & \text{if\quad} iou > T_{pos} \\
        \quad \frac{iou-T_{neg}}{T_{pos}-T_{neg}}, & others
    \end{cases}
\end{equation}
where $iou$ represents the highest Intersection-over-Union ($IoU$) ratio between anchor and all ground-truth boxes, and $T_{neg}$, $T_{pos}$ are pre-defined threshold values. An anchor with $iou$ higher than $T_{pos}$ will be assigned a positive label $1$, or else a negative label $0$ if $iou$ value lower than $T_{neg}$. The rest of anchor boxes will also be assigned a soft label between $0$ and $1$ in accordance with their $IoU$ ratio. Intuitively, the more overlap between the anchor boxes and the ground-truth boxes, the larger the corresponding label value will be. Thus, a series of training samples with continuous label values from $0$ to $1$ are generated.

We define these samples with soft label as semi-positive samples and calculate the number of positive and semi-positive samples generated under different $IoU$ threshold settings in CityPersons\cite{zhang2017citypersons} training set, \emph{i.e.}, \{$0.3$, $0.4$\}, \{$0.4$, $0.5$\} and \{$0.5$, $0.6$\} for $\{T_{neg}, T_{pos}\}$ respectively, which is summarized in Table \ref{table2}. It is clearly that incorporated with the design of soft label, although the difference between $T_{neg}$ and $T_{pos}$ is only set to $0.1$, the number of semi-positive samples generated still exceeds the number of positive samples by a large margin. Strengthening the use of these samples can not only increase the number of samples available, but also enhance the robustness of the model for labels. Figure \ref{Figure-soft} depicts the distribution of labels of the semi-positive samples under different $IoU$ threshold settings, \emph{i.e.}, \{$0.3$, $0.4$\}, \{$0.4$, $0.5$\} and \{$0.5$, $0.6$\} for $\{T_{neg}, T_{pos}\}$. The number of semi-positive samples decreases with the increase of label value naturally. We empirically choose \{$0.4$, $0.5$\} and \{$0.5$, $0.6$\} for the two-step refinement of anchor boxes, which achieves the lowest MR$^{-2}$ in our experiments.

\begin{table}[H]
  \centering
  \caption{\label{table2} The number of positive and semi-positive samples generated under different $IoU$ threshold settings in CityPersons training set. \{$T_{neg}$, $T_{pos}$\} represents the $IoU$ threshold to assign positives and semi-positives defined in Section. \ref{sec-soft}. }
  \vspace{1mm}
  \begin{tabular}{p{1.5cm}<{\centering}| p{1.4cm}<{\centering} | p{2.0cm}<{\centering} | p{1.4cm}<{\centering}}
    \hline
    $T_{neg}$, $T_{pos}$ & \textbf{Positive} & \textbf{Semi-positive} & \textbf{Total} \\
    \hline
    \{$0.3$, $0.4$\} & $122,520$ & $136,383$ & $258,903$ \\
    \hline
    \{$0.4$, $0.5$\} & $56,409$ & $67,334$ & $123,743$ \\
    \hline
    \{$0.5$, $0.6$\} & $29,471$ & $32,954$ & $62,425$ \\
    \hline
  \end{tabular}
  \end{table}

In this way, for classification branch of network, the optimization objective is transformed from binary classification of $0$ and $1$ to a regression process of label learning with value ranging from $0$ to $1$. Therefore, not only more valid positive samples are generated, but also the robustness of detector is strengthened.

\subsection{Center-$IoU$ loss}\label{sec-center}
Intersection-over-Union ($IoU$) is the most popular evaluation metric used in the object detection benchmarks. However, as illustrated in \cite{rezatofighi2019generalized}, there exists a gap between optimizing the commonly used distance losses for regressing the parameters of a bounding box and maximizing this metric value. To alleviate the aforementioned problem, some improved versions such as $GIoU$ loss\cite{rezatofighi2019generalized} and Distance-$IoU$ loss\cite{zheng2019distance} are proposed as the bounding box regression loss functions to replace the commonly used $L_n$-norm loss or $IoU$ loss. In this paper, the $GIoU$ loss is selected as the optimization benchmark, and the definition of this loss function is as follows.

\begin{equation}\label{equa5}
    L_{GIoU}=1-IoU(B_{gt}, B_{pred})+\frac{|C \setminus (B_{gt} \cup B_{pred})|}{|C|},
\end{equation}
where $B_{gt}$ and $B_{pred}$ are the ground-truth and predicted bounding boxes, respectively. $C$ is the smallest rectangular box enclosing both $B_{gt}$ and $B_{pred}$. $IoU(B_{gt}, B_{pred})$ is the $IoU$ ratio of these two bounding boxes. $|C \setminus (B_{gt} \cup B_{pred})|$ represents the volume (area) occupied by $C$ excluding $B_{gt}$ and $B_{pred}$.

\begin{figure}[H]
\centering
\includegraphics[scale=0.55]{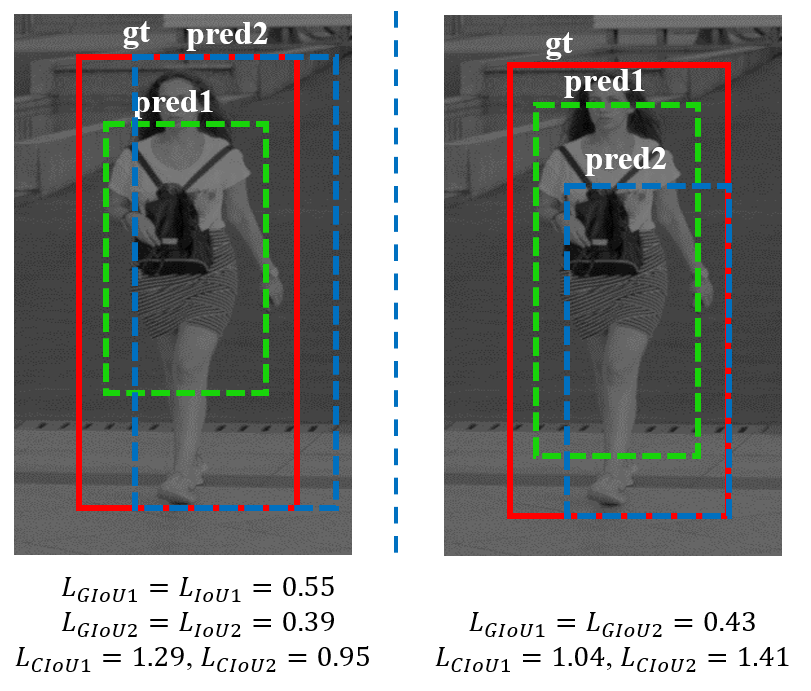}
\caption{Some hard examples for $GIoU$ loss. The box in red represents the ground-truth box, and the boxes in green and blue are two different prediction results.}
\label{figure2}
\end{figure}

Although the $GIoU$ loss has achieved great improvements, some problems still exist. As illustrated in Figure \ref{figure2}, the red rectangular box represents the ground-truth box, the green and blue dotted boxes are two different prediction results. On one hand, the left image depicts that if the predicted box is completely enclosed by the ground-truth box, or these two boxes are in parallel state, the value of term $|C \setminus (B_{gt} \cup B_{pred})|\ / \ |C|$ will fixed to $0$ and the $GIoU$ loss will then degenerate into a common $IoU$ loss. On the other hand, for the right image in Figure \ref{figure2}, we can find that no matter how we move the position of the predicted box, the $GIoU$ loss is always a fixed value on the premise that the predicted box are fully covered by the ground-truth box. As a result, this unconstrained deviation will undoubtedly produce more false negatives for adjacent overlapping pedestrians, especially for occluded pedestrian detection.

To handle these problems, we propose a new bounding box regression loss, named Center-$IoU$ loss. The definition of this loss function is detailed in Equation. \ref{equa6} and \ref{equa7}.

\begin{equation}\label{equa6}
    L_{CIoU}=\text{smooth}_{\text{ln}}(\frac{|C \setminus (B_{gt} \cap B_{pred})|}{|C|}) + \text{smooth}_{\text{L}1}(t_i, t_i^{*}),
\end{equation}
\begin{equation}\label{equa7}
    \text{smooth}_{\text{ln}}(x)=
        \begin{cases}
        -\text{ln}(1-x), & \text{if\quad} x \leq \sigma, \\
        \frac{x-\sigma}{1-\sigma}-\text{ln}(1-\sigma), & \text{if\quad} x > \sigma,
        \end{cases}
\end{equation}
where $\text{smooth}_{\text{ln}}(x)$\cite{wang2018repulsion} is a smoothed $\text{ln}(x)$ function which is continuously differentiable in ($0$, $1$), and $\sigma \in $ [$0$, $1$) is the smooth parameter to adjust the sensitiveness of the regression loss to the outliers. The $IoU$ regression term $|C \setminus (B_{gt} \cap B_{pred})|$ represents the area occupied by $C$ excluding the intersection of $B_{gt}$ and $B_{pred}$. $t_i$ and $t_i^{*}$ are vectors representing the $2$ parameterized center point coordinates of the predicted and ground-truth box, respectively. The distance regression term $\text{smooth}_{\text{L}1}(t_i, t_i^{*})$ is the robust distance loss function defined in \cite{wang2014robust}, which is applied to enforce the predicted box locate closer to the center of ground-truth box, and thus reduce the deviation. As shown in Figure \ref{figure2}, our Center-$IoU$ loss can easily handle the hard examples for $GIoU$ loss, which proves the superiority of our method.

The proposed Center-$IoU$ loss is an organic combination of coordinate regression and $IoU$ regression, which not only retains the good characteristics of $GIoU$ loss, but also solves some defects of it, enabling the regression process more suitable for pedestrian detection.

\subsection{Cosine-NMS}
Non-Maximum Suppression (NMS) is an integral part of the object detection pipeline, which recursively selects the detection box with the maximum score and remove the repeated predictions. However, this greedy NMS algorithm still has some problems. Applying NMS with a low threshold like $0.3$ may increase the miss-rate, especially in crowd scenes. On the contrary, a high threshold like $0.6$ may also increase false positives as many neighboring proposals that are overlapped often have correlated scores. Therefore, for the task of pedestrian detection, the weight function design in NMS needs to meet the following demands to better detect occluded pedestrians.

$1)$ For the detection boxes which are far away from $M$ ($\text{iou}(M, b_i) < N_t$), they have a smaller likelihood of being false positives and they should thus be retained with no penalty imposed, so as to reduce the false negatives caused by non-maximum suppression in the postprocess of occluded pedestrian detection.

$2)$ For the highly overlapped neighboring detection boxes ($\text{iou}(M, b_i) \geq N_t$), they have a higher likelihood of being false positives, the penalty should be higher to quickly reduce the score of these predictions. Furthermore, for the fully overlapped detection boxes ($\text{iou}(M, b_i) = 1$), the classification score of them should be reduced to $0$ since these detections have a high probability of being repeated predictions.

To meet these requirements, we propose a novel Cosine-NMS for the postprocess of predictions based on the design of Soft-NMS\cite{bodla2017soft}. The pruning step of this algorithm can be written as a re-scoring function as follows.
\begin{equation}\label{equa8}
    s_i=
        \begin{cases}
            s_i, & \quad \text{iou}(M, b_i) < N_t, \\
            s_if(\text{iou}(M, b_i)), & \quad \text{iou}(M, b_i) \geq N_t,
        \end{cases}
\end{equation}
where $N_t$ is the pre-defined threshold, $f(\text{iou}(M, b_i))$ is an overlap based weighting function to change the classification score $s_i$ of a box $b_i$ which has a high overlap with $M$. For the detection boxes with $IoU$ ratio lower than $N_t$, they are more likely to be true positives and thus no processing is done to them, otherwise a penalty term will be applied to decay their scores. As for the design of weighting function, we propose a new weighting function based on cosine function to make the NMS algorithm more suitable for pedestrian detection, which is detailed as follows.

\begin{equation}\label{equa9}
    f(\text{iou}(M, b_i))=\text{cos}(\frac{\pi}{2}(\text{iou}(M, b_i)-N_t)/(1-N_t)),
\end{equation}

\begin{figure}[!ht]
\centering
\includegraphics[scale=0.6]{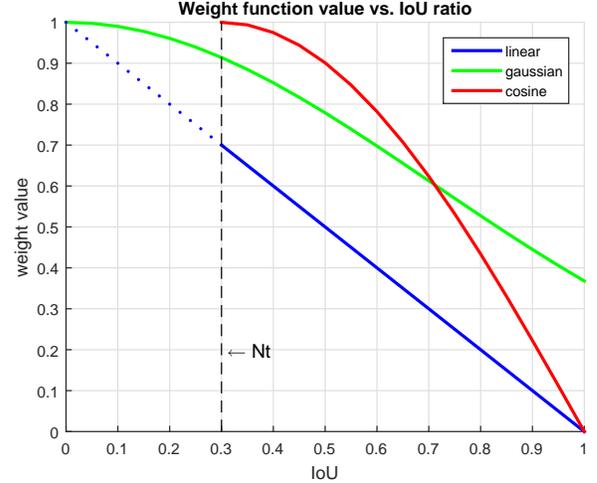}
\setlength{\belowcaptionskip}{0pt}
\setlength{\abovecaptionskip}{0pt}
\caption{The curves of weight value under different forms of weighting function.}
\label{figure3}
\end{figure}

In soft-NMS, either the Linear version $f(\text{iou}(M, b_i))=(1-\text{iou}(M, b_i))$ or Gaussian version $f(\text{iou}(M, b_i))=e^{-\frac{\text{iou}(M, b_i)^2}{\sigma}}$ also decays the scores of detections as an increasing function of overlap with $M$. In order to explain the superiority of the weighting function proposed in this paper, we depict the comparison curves of the three weighting functions in Figure \ref{figure3}. The horizontal axis represents the $IoU$ ratio, and the vertical axis is the corresponding weight value.

The line in blue is the Linear function in Soft-NMS, it is not continuous in terms of overlap with a fixed gradient and a sudden penalty is applied when a NMS threshold of $N_t$ is reached, making the NMS algorithm extremely sensitive to the threshold and the decay of score is not smooth enough for the detections whose $IoU$ value is around $0.3$. Meanwhile, the line in green is the Gaussian function in Soft-NMS, it is continuous in terms of overlap, but the penalty term is introduced to any prediction box overlaps with ground-truth box and is too small for the highly overlapped boxes. Particularly, for the fully overlapped detection boxes ($\text{iou}(M, b_i) = 1$), the decayed classification score of them is still high and cannot reach $0$, which may cause false positives in the postprocess of detections. Therefore, these two weighting functions cannot meet the requirements of pedestrian detection, which need further optimization.

For comparison, the line in red is the proposed Cosine penalty function for NMS, it is continuous in terms of overlap when the $IoU$ ratio is higher than threshold $N_t$. For the detection boxes with $IoU$ ratio lower than $N_t$, they are more likely to be true positives and thus no processing is done to them, otherwise a penalty term will be applied to decay their scores. Besides, it is also clear that scores for detection boxes which have a higher overlap with $M$ will be decayed more with a higher penalty, as they have a higher likelihood of being false positives. Particularly, for the fully overlapped detection boxes ($\text{iou}(M, b_i) = 1$), the decayed classification score of them will be further decayed to $0$ to reduce false positives of the detector. Notably, the proposed Cosine-NMS has no additional parameters and can be directly applied to the post-processing of any pedestrian detectors.

\subsection{Adaptive anchor matching}
\begin{figure}[!ht]
  \centering
  \subfigure[Image]{
  \label{image}
  \begin{minipage}{3.9cm}
  \centering
    \includegraphics[width=4.2cm]{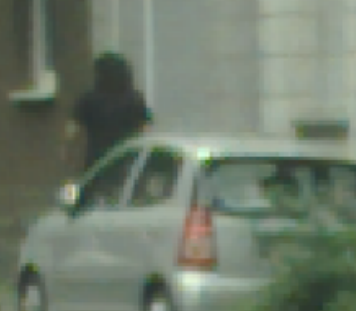}
  \end{minipage}
  }
  \subfigure[Bounding box anno]{
  \label{anno}
  \begin{minipage}{4.1cm}
  \centering
    \includegraphics[width=4.2cm]{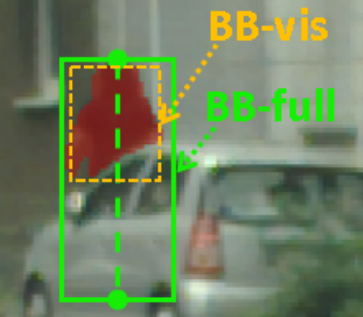}
  \end{minipage}
  }
  \caption{An example of the visible part and full body of pedestrian annotation in CityPersons dataset.}
  \label{figure4}
\end{figure}

In common paradigm of anchor matching, the $IoU$ ratios are usually calculated between pre-defined anchor boxes and full ground-truth bounding boxes based on the jaccard overlap. However, as shown in Figure \ref{figure4}, for the occluded pedestrian, the visible bounding box for each instance is only part of full bounding box, which contains a lot of redundant background information and makes it invalid to directly match the anchor to the full bounding box. To obtain a better result, we propose adaptive anchor matching to optimize this process as follows.

$1)$ For ground-truth boxes with visible ratio $R_{vis}$ lower than threshold $T_{vis}$, apply the visible bounding boxes to match the anchor boxes to reduce the interference of background information. The visible ratio $R_{vis}$ is defined as the ratio of the visible bounding box and full bounding box, \emph{i.e.} $area(B_\text{vis}) / area(B_\text{full})$.

$2)$ For ground-truth boxes with visible ratio $R_{vis}$ higher than threshold $T_{vis}$, directly apply the full bounding boxes to match the anchor boxes to make full use of the context information to assist pedestrian detection.

$3)$ After the anchor matching, the regression target of anchor box is still the full bounding box of ground-truth in both cases mentioned above. The threshold $T_{vis}$ is fixed to $0.5$ in all datasets, which presents the best MR$^{-2}$ in our experiments.

\subsection{Other implementations}
\textbf{Training.} Anchors are assigned as positives $S_{+}$ if the $IoUs$ with any ground truth are above a threshold $T_{pos}$, and negatives $S_{-}$ if the $IoUs$ lower than a threshold $T_{neg}$. Those anchors with $IoU$ in [$T_{neg}$, $T_{pos}$) are also assigned as semi-positives $S_{*}$ with label value ranging from $0$ to $1$ and sent to the training objective as well, which is detailed in Section. \ref{sec-soft}. We assign different $IoU$ threshold sets \{$T_{neg}$, $T_{pos}$\} for different regression steps with \{0.4, 0.5\} for step $1$ and \{0.5, 0.6\} for step $2$.

At each regression step $t$, the convolutional predictor is optimized by a multi-task loss function combining two objectives, which is illustrated as follows.

\begin{small}
\begin{equation}\label{equa_10}
    L(p, x)= \frac{1}{N_{cls}} \sum_{i=1} L_{cls}(p_i, p_i^*) + \frac{\lambda}{N_{reg}} \sum_{i=1} [p_i^*>0]L_{reg}(x_i, x_i^*),
\end{equation}
\end{small}

\noindent
where $i$ is the index of an anchor, $p_i$ and $p_i^*$ are the predicted probability and ground-truth label of anchor $i$. $x_i$ is a vector representing the $4$ parameterized coordinates of the predicted locations of pedestrian, while $x_i^*$ is the corresponding ground-truth box parameters associated with a non-negative anchor. The regression loss $L_{reg}$ is the proposed Center-$IoU$ loss detailed in Section. \ref{sec-center}. $[p_i^*>0]L_{reg}$ means the bounding box regression loss is activated only for non-negative anchors. The classification loss $L_{cls}(p_i, p_i^*)$ is an optimized Focal Loss\cite{lin2017focal} for positive, negative and semi-positive samples, which is formulated as:

\begin{small}
\begin{equation}\label{equa_11}
    \begin{aligned}
    L_{cls}(p_i, p_i^*) = &-\alpha \sum_{i\in S_{+}}(1-p_i)^{\gamma}\text{log}(p_i) - \beta \sum_{i\in S_{*}}(p_i^*)^{\gamma} \text{log}(p_i) \\
                        & -(1-\alpha) \sum_{i\in S_{-}}p_i^\gamma \text{log}(1-p_i),
    \end{aligned}
\end{equation}
\end{small}

\noindent
where $\alpha$ and $\gamma$ are the focusing parameters, which are set to $0.25$ and $2$ as suggested in \cite{lin2017focal}. The first and third terms in loss function is the traditional Focal Loss for binary classification of positive and negative samples. While the second term is the specially designed loss function for semi-positive samples with balancing parameter $\beta$ fixed to $0.1$, which assigns greater weights to the samples with higher label value since these samples are more important for model training to alleviate the positive-negative imbalance problem.

The classification loss and regression loss are then normalized by $N_{cls}$ and $N_{reg}$, and further weighted by balancing parameter $\lambda$. In our implementation, the $cls$ term is normalized by the number of positive, negative and semi-positive anchors, and the $reg$ term is normalized by the number of non-negative anchors.

\textbf{Data augmentation.} To increase the robustness of training data, each training image is sequentially processed by color distortion and horizontal flipping with a probability of $0.5$. For CityPersons\cite{zhang2017citypersons} and Caltech\cite{dollar2011pedestrian} datasets, we randomly crop a patch with the size of [$0.3$, $1.0$] of the original image and further resize it so that the shorter side has $N$ pixels ($N=640$ for CityPersons and $N=336$ for Caltech), while keeping the aspect ratio of the image. As for CrowdHuman\cite{shao2018crowdhuman} dataset, since the images of this dataset are collected from Web sites with various sizes. We randomly crop a square patch with the size of [$0.3$, $1.0$] of the short size of the original image, and then resize it to $832\times832$ pixels as the final training image in our experiments.

\textbf{Inference.} SADet simply involves feeding forward an image through the network. For each level, we get the regressed anchor boxes from the final predictor and hybrid confidence scores from all predictors. We first filter out most boxes by a confidence threshold of $0.05$ and keep the top
$1000$ boxes before applying NMS, then all remaining boxes are merged with the proposed Cosine Non-Maximum Suppression (Cosine-NMS) with a threshold of $0.3$, and the top $150$ boxes will be selected as the output.

\begin{table*}
  \centering
  \caption{\label{table3}Comparison with the state-of-the-art methods on the CityPersons validation set. All models are trained on the training set. Detection results tested on the original image size ($1024\times2048$ on CityPersons) are reported. In the column of Framework, One-stage and Two-stage are both the anchor-based methods, and Anchor-free means the anchor-free method. SADet-\emph{n}step represents the model with \emph{n} steps refinement to anchor boxes.}
  \vspace{1mm}
  \begin{tabular}{p{3.9cm}<{\centering} |p{1.8cm}<{\centering} |p{1.6cm}<{\centering} |p{1.6cm}<{\centering} |p{1.2cm}<{\centering} p{1.2cm}<{\centering} p{1.2cm}<{\centering} |p{1.5cm}<{\centering}}
    \hline
    Method & Framework & Backbone & \textbf{Reasonable} & Bare & Partial & Heavy & Test Time \\
    \hline
    \hline
    Adapted Faster RCNN\cite{zhang2017citypersons} & Two-stage & VGG-16 & 15.4 & - & - & - & - \\
    OR-CNN\cite{zhang2018occlusion} & Two-stage & VGG-16 & 12.8 & 6.7 & 15.3 & 55.7 & - \\
    Adaptive-NMS\cite{liu2019adaptive} & Two-stage & VGG-16 & 11.9 & 6.2 & 12.6 & 55.2 & - \\
    TLL+MRF\cite{song2018small} & Two-stage & ResNet-50 & 14.4 & 9.2 & 15.9 & 52.0 & - \\
    Repulsion Loss\cite{wang2018repulsion} & Two-stage & ResNet-50 & 13.2 & 7.6 & 16.8 & 56.9 & - \\
    ALFNet-2s\cite{liu2018learning} & One-stage & ResNet-50 & 12.0 & 8.4 & 11.4 & 51.9 & 0.27s/img\\
    CSP(w/o offset)\cite{liu2019high} & Anchor-free & ResNet-50 & 11.4 & 8.1 & 10.8 & 49.9 & 0.33s/img\\
    CSP(with offset)\cite{liu2019high} & Anchor-free & ResNet-50 & 11.0 & 7.3 & 10.4 & \textcolor{red}{49.3} & 0.33s/img\\
    APD\cite{zhang2019Attribute} & Anchor-free & ResNet-50 & 10.6 & 7.1 & \textcolor{red}{9.5} & 49.8 & - \\
    \hline
    SADet-1step & One-stage & ResNet-50 & 11.5 & 6.7 & 10.7 & 56.4 & \textcolor{red}{0.24s/img} \\
    SADet-2step & One-stage & ResNet-50 & \textcolor{red}{\textbf{9.7}} & \textcolor{red}{5.7} & 9.8 & 52.8 & 0.26s/img\\
    \hline

  \end{tabular}
\end{table*}

\section{Experiments}
\subsection{Experimental setup}
Our method is implemented in the Keras\cite{chollet2015keras} library, with $2$ GTX TITAN X GPUs for model training. A mini-batch contains $15$ images per GPU. The Adam\cite{kingma2014adam} solver is applied. For CityPersons\cite{zhang2017citypersons}, the backbone network is pre-trained on ImageNet\cite{deng2009imagenet} and all added layers are randomly initialized with the ``$xavier$'' method\cite{glorot2010understanding}. The network is totally trained for $120$ epoches, with the initial learning rate of $1e^{-3}$ and decreased by a factor of $10$ after $60$ and $80$ epoches. For Caltech\cite{dollar2011pedestrian}, we also include experiments with the model initialized from CityPersons as done in \cite{liu2018learning} and totally trained for $100$ epoches with the learning rate of $1e^{-5}$. For CrowdHuman\cite{shao2018crowdhuman}, the network is totally trained for $140$ epoches with the base learning rate $1e^{-3}$ and divided by $10$ after $80$ and $100$ epoches. The backbone network is ResNet-$50$ unless otherwise stated.

\subsection{CityPersons dataset}
The CityPersons\cite{zhang2017citypersons} dataset is built upon the semantic segmentation dataset Cityscapes to provide a new dataset of interest for pedestrian detection, which contains $5,000$ images ($2,975$ for training, $500$ for validation, and $1,525$ for testing) with about $35,000$ manually annotated persons plus $13,000$ ignore region annotations. Both the bounding boxes and visible parts of pedestrians are provided and there are approximately $7$ pedestrians in average per image.

Following the evaluation protocol in CityPersons, we train our detector on the training set, and evaluate it on the validation sets. The log miss rate averaged over the false positive per image (FPPI) range of [$10^{-2}$, 1] (MR$^{-2}$) is used to measure the detection performance (lower score indicates better performance).

\begin{table}[H]
  \centering
  \caption{\label{table4}Ablative results on CityPersons validation set. MR$^{-2}$ is used to compare the performance of detectors (lower score indicates better performance). The top result is highlighted in red. }
  \vspace{1mm}
  \begin{tabular}{p{1.9cm}<{\centering}| p{0.7cm}<{\centering} p{0.7cm}<{\centering} p{0.7cm}<{\centering} p{0.7cm}<{\centering} p{0.7cm}<{\centering}}
    \hline
    \textbf{Component} & \multicolumn{5}{c}{\textbf{SADet}}\\
    \hline
    Center-$IoU$ & & $\surd$ & $\surd$ & $\surd$ & $\surd$ \\
    Soft Label & & & $\surd$ & $\surd$ &$\surd$ \\
    Cosine-NMS & & & & $\surd$ & $\surd$ \\
    Adaptive & & & & & $\surd$ \\
    \hline
    \textbf{Reasonable} & \textbf{16.0} & \textbf{15.2} & \textbf{13.4} & \textbf{12.9} & \textcolor{red}{\textbf{11.5}} \\
    Bare & 11.3 & 10.0 & 8.7 & 8.2 & \textcolor{red}{6.7}  \\
    Partial & 14.8 & 14.2 & 12.9 & 12.1 & \textcolor{red}{10.7} \\
    Heavy & 52.5 & 52.0 & 51.7 & \textcolor{red}{51.5} & 56.4 \\
    \hline
  \end{tabular}
  \end{table}

\subsubsection{Ablation study}
To have a better understanding of how each proposed component affects the final performance, we construct four variants and evaluate them on CityPersons validation dataset, shown in Table \ref{table4}. Our baseline model is the same as ALFNet\cite{liu2018learning}, with smooth $L_1$ loss for bounding box regression and Focal Loss for classification.

\textbf{Center-IoU Loss.} Firstly, the smooth $L_1$ loss for bounding box regression applied in ALFNet is replaced with the novel Center-$IoU$ loss, which helps promote the accuracy of localization and contributes to the reduce of MR$^{-2}$ by $0.8\%$ on \emph{Reasonable} subset. Besides, the improvement still holds for pedestrians with different degrees of occlusion, including \emph{Bare} (occlusion $ \le 10\%$), \emph{Partial} ($10\% < $ occlusion $ < 35\%$) and \emph{Heavy} ($35\% \ge $ occlusion) subsets. To better demonstrate the superiority of the Center-$IoU$ loss, we further train the model using different bounding box regression losses, including smooth $L_1$\cite{liu2016ssd}, $IoU$\cite{kosub2019note}, $GIoU$\cite{rezatofighi2019generalized} and Distance-$IoU$\cite{zheng2019distance} losses, as well as our proposed Center-$IoU$ loss, which is detailed in Table \ref{table-center}, where $L_1$ represents the smooth $L_1$ loss, $L_{IoU}$, $L_{GIoU}$ and $L_{DIoU}$ are $IoU$, $GIoU$ and Distance-$IoU$ losses respectively, $L_{CIoU}$ is our Center-$IoU$ loss. Notably, for a fair comparison, we ablate all the other unrelated modules, and directly conduct the experiments on our base model. As shown in Table \ref{table-center}, the proposed Center-$IoU$ loss surpasses all the other losses by a large margin, achieving obvious improvements of $0.8\%$, $0.4\%$, $0.7\%$ and $0.2\%$ MR$^{-2}$ compared with smooth $L_1$, $IoU$, $GIoU$ and Distance-$IoU$ losses, respectively.

\begin{table}[H]
  \centering
  \caption{\label{table-center} Comparison between the performance of the base model trained using smooth $L_1$ loss as well as $L_{IoU}$, $L_{GIoU}$, $L_{DIoU}$ and our $L_{CIoU}$ losses. The results are reported on the validation set of CityPersons.}
  \vspace{1mm}
  \begin{tabular}{p{1.2cm}<{\centering}| p{0.8cm}<{\centering}  p{0.8cm}<{\centering} p{0.8cm}<{\centering} p{0.8cm}<{\centering} p{0.8cm}<{\centering} }
    \hline
    Loss & $L_1$ & $L_{IoU}$ & $L_{GIoU}$ & $L_{DIoU}$ & $L_{CIoU}$ \\
    \hline
    MR$^{-2}(\%)$ & $16.0$ & $15.6$ & $15.9$ & $15.4$ & \textbf{15.2} \\
    \hline
  \end{tabular}
  \end{table}

\textbf{Soft Label Design.} Secondly, we incorporate the soft label design to sample generation process to fully utilize all valid samples and smooth the predictions of the model. The comparison between the third and fourth columns in Table \ref{table4} indicates that our soft label design effectively improves the performance, especially for \emph{Reasonable} subset with MR$^{-2}$ reduced from $15.2\%$ to $13.4\%$.

\begin{table}[H]
  \centering
  \caption{\label{table-cosine} Comparison results of the base model using different Non-Maximum Suppression algorithms, including greedy-NMS, Soft-NMS and Cosine-NMS.}
  \vspace{1mm}
  \begin{tabular}{p{1.2cm}<{\centering}| p{0.8cm}<{\centering}  p{0.8cm}<{\centering} p{0.85cm}<{\centering} p{0.8cm}<{\centering} }
    \hline
    NMS & greedy & soft-$l$ & soft-$g$ & cosine\\
    \hline
    MR$^{-2}(\%)$ & $16.0$ & $15.7$ & $16.1$ & \textbf{15.4} \\
    \hline
  \end{tabular}
  \end{table}

\textbf{Cosine-NMS.} Thirdly, we substitute the proposed Cosine-NMS for the commonly used greedy-NMS to make the Non-Maximum Suppression algorithm more suitable for pedestrian detection. The result in Table \ref{table4} validates the effectiveness of Cosine-NMS, with MR$^{-2}$ decreased by $0.5\%$, $0.5\%$, $0.8\%$ and $0.2\%$ on \emph{Reasonable}, \emph{Bare}, \emph{Partial} and \emph{Heavy} subsets. In Table \ref{table-cosine}, we compare our Cosine-NMS with traditional NMS and Soft-NMS on CityPersons dataset, where $greedy$ means the greedy-NMS, soft-$l$ and soft-$g$ are the Linear and Gaussian versions of Soft-NMS, $cosine$ represents the proposed Cosine-NMS. Similarly, we also ablate all the other unrelated modules, and conduct the experiments on our base model directly. Table \ref{table-cosine} shows that our Cosine-NMS method achieves the best MR$^{-2}$ of $15.4\%$ on the base model, outperforming all the other Non-Maximum Suppression algorithms, which demonstrates the effectiveness of the proposed method in pedestrian detection.

\begin{table*}
  \centering
  \caption{\label{table5} Comparisons of MR$^{-2}$ and running time on caltech dataset. The time of CompACT-Deep and RPN+BF are reported in\cite{zhang2016faster}, and that of SA-FastRCNN and F-DNN are reported in. MR$^{-2}$ is based on the new annotations\cite{zhang2017citypersons}. The original image size on Caltech is $640\times480$. SADet-\emph{n}step represents the model with \emph{n} steps refinement to anchor boxes.}
  \vspace{1mm}
  \begin{tabular}{p{2.9cm}<{\centering}| p{1.2cm}<{\centering} | p{2.2cm}<{\centering} | p{1.8cm}<{\centering} | p{1.5cm}<{\centering} | p{1.5cm}<{\centering}}
    \hline
    \multirow{2}{*}{Method} & \multirow{2}{*}{Scale} & \multirow{2}{*}{Hardware} & \multirow{2}{*}{Test time} & \multicolumn{2}{c}{MR$^{-2}$} \\
    \cline{5-6}
      &   &   &   & IoU=0.5 & IoU=0.75 \\
    \hline
    \hline
    CompACT-Deep\cite{cai2015learning} & $\times 1.0$ & Tesla K40 & $0.5$ s/img & $9.2$ & $59.0$ \\
    RPN+BF\cite{zhang2016faster} & $\times 1.5$ & Tesla K40 & $0.5$ s/img & $7.3$ & $57.8$ \\
    SA-FastRCNN\cite{li2017scale} & $\times 1.7$ & Titan X & $0.59$ s/img & $7.4$ & $55.5$ \\
    F-DNN\cite{du2017fused} & $\times 1.0$ & Titan X & $0.16$ s/img & $6.9$ & $59.8$ \\
    Zhang \emph{et al.}\cite{zhang2017citypersons} & $\times 1.0$ & $-$ & $-$ & $5.8$ & $30.6$ \\
    Repulsion Loss\cite{wang2018repulsion} & $\times 1.0$ & $-$ & $-$ & $4.0$ & $23.0$ \\
    ALFNet\cite{liu2018learning} & $\times 1.0$ & GTX $1080$Ti & $0.05$ s/img & $4.5$ & $18.6$ \\
    \hline
    SADet-1step & $\times 1.0$ & GTX $1080$Ti & $0.05$ s/img & $5.1$ & $19.4$ \\
    SADet-2step & $\times 1.0$ & GTX $1080$Ti & $0.05$ s/img & \textbf{3.8} & \textbf{16.3} \\
    \hline
  \end{tabular}
  \end{table*}

\textbf{Adaptive Anchor Matching.} The last contribution of SADet is the proposed adaptive anchor matching, which deals with the the problem of poor matching between anchor boxes and ground-truth boxes in crowded scenes. As reported in Table \ref{table4}, the improvements on
\emph{Reasonable}, \emph{Bare}, \emph{Partial} subsets are $1.4\%$, $1.5\%$ and $1.4\%$ respectively. Meanwhile, the performance under \emph{Heavy} subset deteriorates to some extent, since the pedestrians under heavy occlusion are matched with anchors with small sizes, making it harder to regress to the full bounding boxes. Overall, this module is still an excellent selection for the improvement of pedestrian detection, especially for the scenes with non-heavy occlusion.

\subsubsection{Evaluation results}
We compare the proposed SADet with the state-of-the-art detectors on CityPersons in Table \ref{table3}. Detection results tested on the original image size are presented. Notably, SADet-$n$step represents the model with $n$ steps refinement to anchor boxes.

Without any additional supervision like semantic labels or auxiliary regression loss, our SADet achieves state-of-the-art results on the validation set of CityPersons by reducing $0.9\%$ MR$^{-2}$ with $\times 1$ scale, surpassing all published anchor-based and anchor-free methods, which demonstrates the superiority of the proposed method in pedestrian detection. Furthermore, following the strategy in \cite{wang2018repulsion}, we also test our method on three subsets with different occlusion levels, \emph{i.e.} \emph{Bare}, \emph{Partial} and \emph{Heavy}. Our method also presents the best or comparable performance in term of different levels of occlusions, demonstrating the self-contained ability of our method to handle occlusion issues in crowded scenes. In addition, we can further conclude from the table that the performance of the anchor-based detector is better on the subsets with low occlusion, \emph{i.e.}, \emph{Reasonable} and \emph{Bare}, while the anchor-free method is more capable of detecting occluded pedestrians with higher performance on \emph{Partial} and \emph{Heavy} subsets. Some qualitative results are depicted in Figure. \ref{figure-citypersons}, threshold is set to $0.4$.

Furthermore, these optimization strategies proposed in this paper are systematic improvements to the detection pipeline of one-stage pedestrian detector, which can be also applied to any anchor-based pedestrian detector, including one-stage and two-stage methods. To validate the generalization capacity and effectiveness of our method, we further conduct the ablative experiments based on different baselines, including one-stage detector SSD\cite{liu2016ssd} and two-stage detector Faster R-CNN\cite{ren2015faster}. For SSD detector, we only adjust the aspect ratio of anchors to $0.41$, and other settings are consistent with the original algorithm. Meanwhile, for Faster R-CNN detector, we follow the adjustments of Adaptive Faster R-CNN\cite{zhang2016faster} to make the algorithm more suitable for pedestrian detection. Then the optimization strategies proposed in this paper are further applied to the RPN module, and the Fast R-CNN module remains unchanged. In the experiment, both these two detectors apply ResNet-$50$ as the backbone network, and the baselines of SSD and Faster R-CNN are $17.2\%$ and $15.9\%$ MR$^{-2}$, respectively. Detection results tested on the original image size ($1024\times2048$ on CityPersons) are reported on Table \ref{table-add}. By sequentially introducing the proposed four modules to the base models, we receive significant improvements with MR$^{-2}$ decreases sharply on different detectors, \emph{i.e.}, $3.8\%$ for SSD and $3.4\%$ for Faster R-CNN, which fully proves the generalization capacity and effectiveness of our method.

\begin{table}[H]
  \centering
  \caption{\label{table-add}Ablative results of different baselines on CityPersons validation set. MR$^{-2}$ is used to compare the performance of detectors (lower score indicates better performance).}
  \vspace{1mm}
  \begin{tabular}{p{1.9cm}<{\centering}| p{0.7cm}<{\centering} p{0.7cm}<{\centering} p{0.7cm}<{\centering} p{0.7cm}<{\centering} p{0.7cm}<{\centering}}
    \hline
    \textbf{Component} & \multicolumn{5}{c}{\textbf{Result}}\\
    \hline
    Center-$IoU$ & & $\surd$ & $\surd$ & $\surd$ & $\surd$ \\
    Soft Label & & & $\surd$ & $\surd$ &$\surd$ \\
    Cosine-NMS & & & & $\surd$ & $\surd$ \\
    Adaptive & & & & & $\surd$ \\
    \hline
    SSD & 17.2 & 16.5 & 15.8 & 15.2 & \textbf{13.4}  \\
    Faster R-CNN & 15.9 & 15.3 & 14.6 & 14.0 & \textbf{12.5}  \\
    \hline
  \end{tabular}
  \end{table}

\subsection{Caltech dataset}
The Caltech\cite{dollar2011pedestrian} is one of the most popular and challenging datasets for pedestrian detection, which comes from approximately $10$ hours $30$ Hz VGA video recorded by a car traversing the streets in the greater Los Angeles metropolitan area. We use the new high quality annotations provided by \cite{zhang2017citypersons} to evaluate the proposed method. The training and testing sets contains $42,782$ and $4,024$ frames, respectively. Following \cite{dollar2011pedestrian}, the log-average miss rate over $9$ points ranging from $10^{-2}$ to $10^0$ FPPI is used to evaluate the performance of the detector.

\begin{figure}[!ht]
\centering
\includegraphics[scale=0.55]{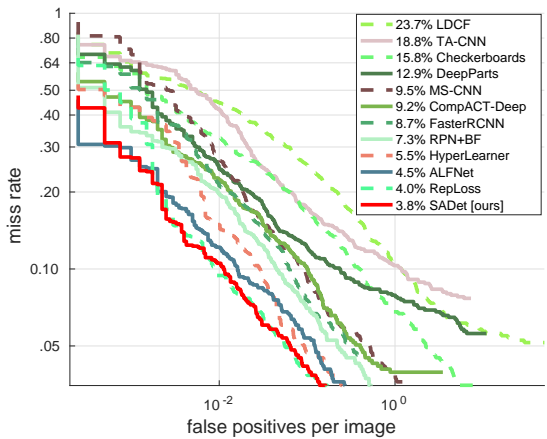}
\setlength{\belowcaptionskip}{0pt}
\setlength{\abovecaptionskip}{0pt}
\caption{Comparisons with the state-of-the-art methods on the Caltech dataset. The scores in the legend are the MR$^{-2}$ scores of the corresponding methods.}
\label{figure5}
\end{figure}

\begin{figure*}
\begin{center}
\includegraphics[scale=0.6]{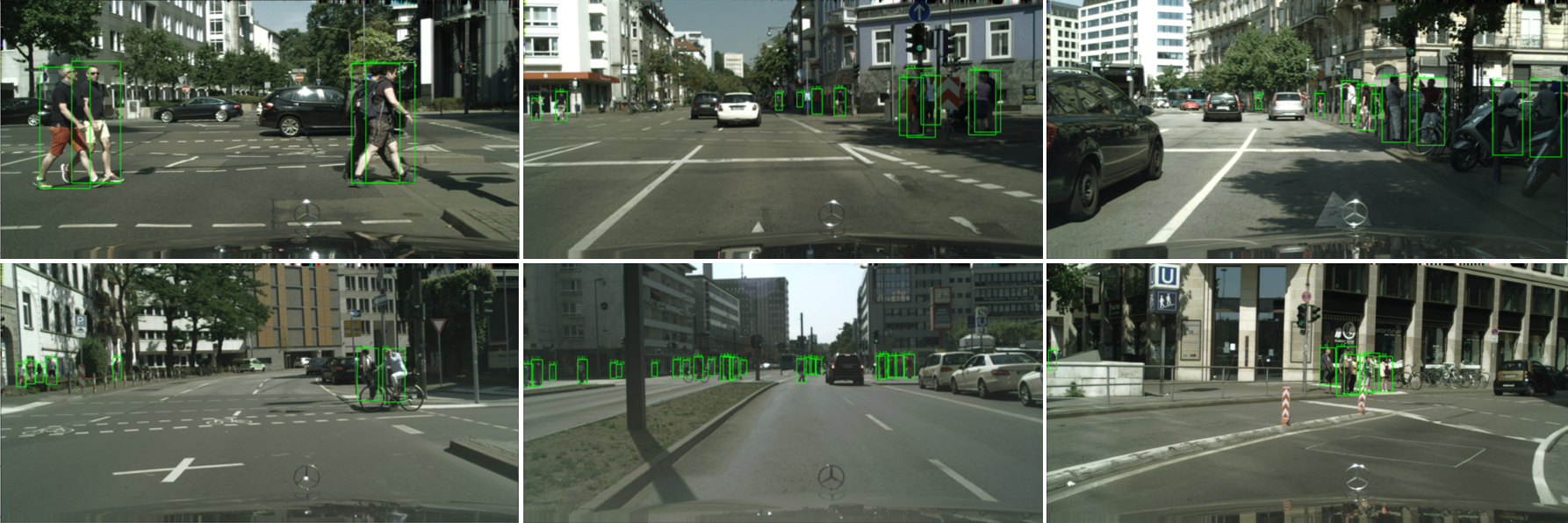}
\end{center}
\caption{Qualitative results on CityPersons pedestrian detection dataset, please zoom in to see some small detections.}
\label{figure-citypersons}
\end{figure*}

We directly fine-tune the detection models pre-trained on CityPersons\cite{zhang2017citypersons} of the proposed SADet on the training set in Caltech. Similar to \cite{wang2018repulsion}, we evaluate the SADet on the \emph{Reasonable} subset of the Caltech dataset, and compare it to other state-of-the-art methods in Figure \ref{figure5}. Notably, the \emph{Reasonable} subset (occlusion $<35\%$) only includes the pedestrians with at least $50$ pixels tall, which is widely used to evaluate the pedestrian detectors.

\begin{figure*}
\begin{center}
\includegraphics[scale=0.563]{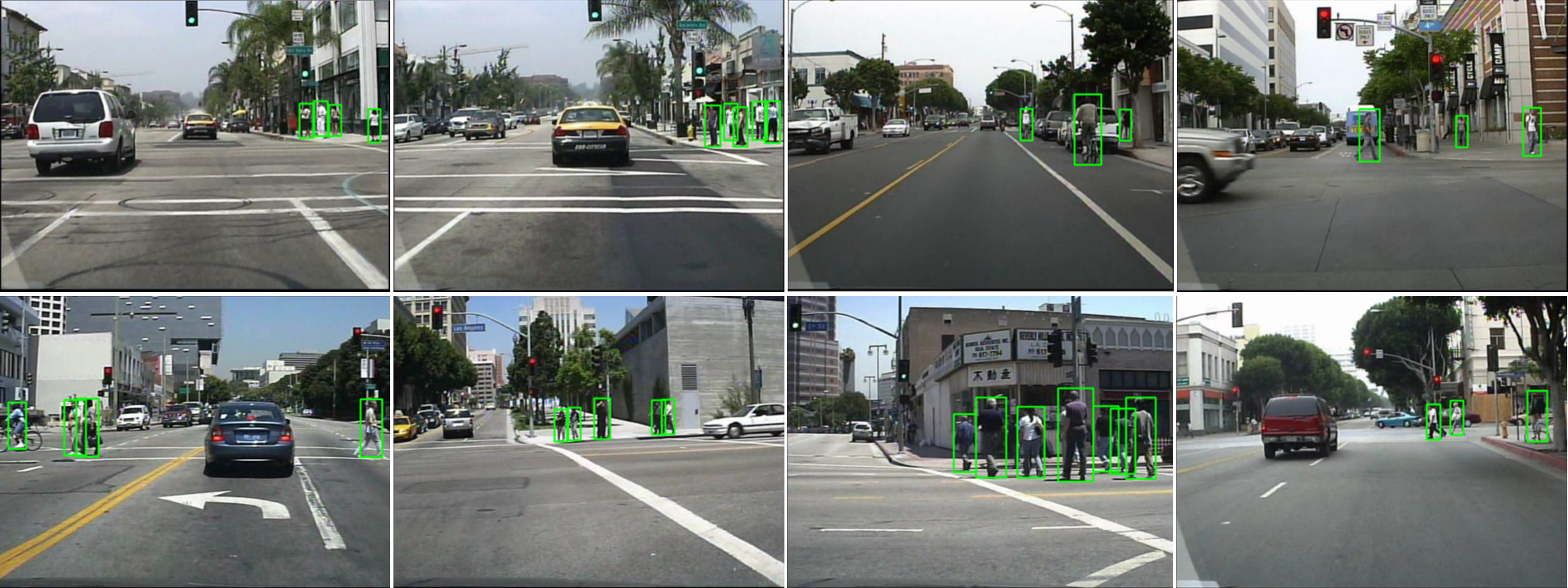}
\end{center}
\caption{Qualitative results on Caltech pedestrian detection dataset, please zoom in to see some small detections.}
\label{figure-caltech}
\end{figure*}

As shown in Figure \ref{figure5}, the proposed SADet performs competitively with the state-of-the-art result\cite{yang2015convolutional, tian2015pedestrian, mao2017can, tian2015deep, daniel2016semantic, liu2018learning, wang2018repulsion} on the Caltech dataset, surpassing all one-stage and two-stage detectors. Besides, the inference time of different methods on Caltech are also reported in Table \ref{table5}, our SADet achieves a MR$^{-2}$ of $3.8\%$ with real-time speed of $20$ FPS for VGA-resolution images ($640\times480$), which presents the effectiveness of our method in pedestrian detection. Some qualitative results are depicted in Figure. \ref{figure-caltech}, threshold is set to $0.4$.

\subsection{CrowdHuman dataset}
Given that the size of CityPersons\cite{zhang2017citypersons} and Caltech\cite{dollar2011pedestrian} datasets are not particularly large, we further carry out experiments on the newly released CrowdHuman\cite{shao2018crowdhuman} dataset to validate the generalization capacity of the proposed method.

\begin{table}[H]
  \centering
  \caption{\label{table6} Volume and density of training sets in different human detection datasets.}
  \vspace{1mm}
  \begin{tabular}{p{2.1cm}<{\centering}| p{1.6cm}<{\centering} p{1.0cm}<{\centering} p{1.8cm}<{\centering}}
    \hline
    \textbf{Dataset} & \textbf{CityPersons} & \textbf{Caltech} & \textbf{CrowdHuman} \\
    \hline
    images & $2,975$ & $\textbf{42,782}$ & $15,000$ \\
    persons & $19,238$ & $13,674$ & $\textbf{339,565}$ \\
    ignore regions & $6,768$ & $50,363$ & $\textbf{99,267}$ \\
    person/image & $6.47$ & $0.32$ & $\textbf{22.64}$ \\
    \hline
  \end{tabular}
  \end{table}

As illustrated in Table \ref{table6}, the CrowdHuman is a large and rich-annotated human detection dataset, which contains $15,000$, $4,370$ and $5,000$ images collected from the Internet for training, validation and testing respectively. The number is more than $10\times$ boosted compared with previous challenging pedestrian detection dataset like CityPersons. The total number of persons is also noticeably larger than the others with $\sim340k$ person and $\sim99k$ ignore region annotations in the CrowdHuman training subset.We follow the evaluation protocol in CityPersons, denoted as MR$^{-2}$. All the experiments are trained in the CrowdHuman training set and evaluated in the validation set.

To reduce the influence of irrelevant variables on the experiments, we still use the same backbone and settings of design parameters applied in the experiments of CityPersons\cite{zhang2017citypersons} and Caltech\cite{dollar2011pedestrian}, such as $0.41$ anchor ratio and $4$ detection layers. Besides, we follow the settings of \cite{shao2018crowdhuman} to improve the baseline result by employing the Feature Pyramid Network (FPN) in our model. As a result, our base model finally achieves a MR$^{-2}$ of $67.64\%$ on CrowdHuman dataset.

\begin{table}[H]
  \centering
  \caption{\label{table7}Ablative results on CrowdHuman validation set. MR$^{-2}$ is used to compare the performance of detectors (lower score indicates better performance).}
  \vspace{1mm}
  \begin{tabular}{p{1.9cm}<{\centering}| p{0.7cm}<{\centering} p{0.7cm}<{\centering} p{0.7cm}<{\centering} p{0.7cm}<{\centering} p{0.7cm}<{\centering}}
    \hline
    \textbf{Component} & \multicolumn{5}{c}{\textbf{SADet}}\\
    \hline
    Center-$IoU$ & & $\surd$ & $\surd$ & $\surd$ & $\surd$ \\
    Soft Label & & & $\surd$ & $\surd$ &$\surd$ \\
    Cosine-NMS & & & & $\surd$ & $\surd$ \\
    Visibility & & & & & $\surd$ \\
    \hline
    MR$^{-2}$ & 67.64 & 66.22 & 64.78 & 64.32 & \textbf{62.96} \\
    \hline
  \end{tabular}
  \end{table}

Table \ref{table7} presents the ablative results of proposed method under the same settings of ablative experiments conducted in CityPersons dataset. By sequentially introducing the proposed four modules to our base model, we receive significant improvements with MR$^{-2}$ decreased by $1.42\%$, $1.44\%$, $0.46\%$ and $1.36\%$ respectively, achieving $62.96\%$ MR$^{-2}$ on CrowdHuman dataset with one step regression, which shows the effectiveness of the proposed method.

\begin{table}[H]
  \centering
  \caption{\label{table8} Evaluation of full body detections on the CrowdHuman validation set.}
  \vspace{1mm}
  \begin{tabular}{p{3.2cm}<{\centering}| p{1.1cm}<{\centering} p{1.1cm}<{\centering} p{1.1cm}<{\centering}}
    \hline
    Method & MR$^{-2}$ & Recall & AP \\
    \hline
    RetinaNet\cite{lin2017focal} & 63.33 & 93.80 & 80.83 \\
    ALFNet\cite{liu2018learning} & 64.37 & 92.10 & 80.13 \\
    RFBNet\cite{liu2019adaptive} & 65.22 & 94.13 & 78.33 \\
    RFBNet-adaptive\cite{liu2019adaptive} & 63.03 & 94.77 & 79.67 \\
    \hline
    \hline
    Baseline & 67.64 & 90.14 & 77.23 \\
    SADet-1step & 62.96 & 94.22 & 80.01 \\
    SADet-2step & \textbf{60.13} & \textbf{95.02} & \textbf{82.14} \\
    \hline
  \end{tabular}
  \end{table}

\begin{figure*}
\begin{center}
\includegraphics[scale=0.649]{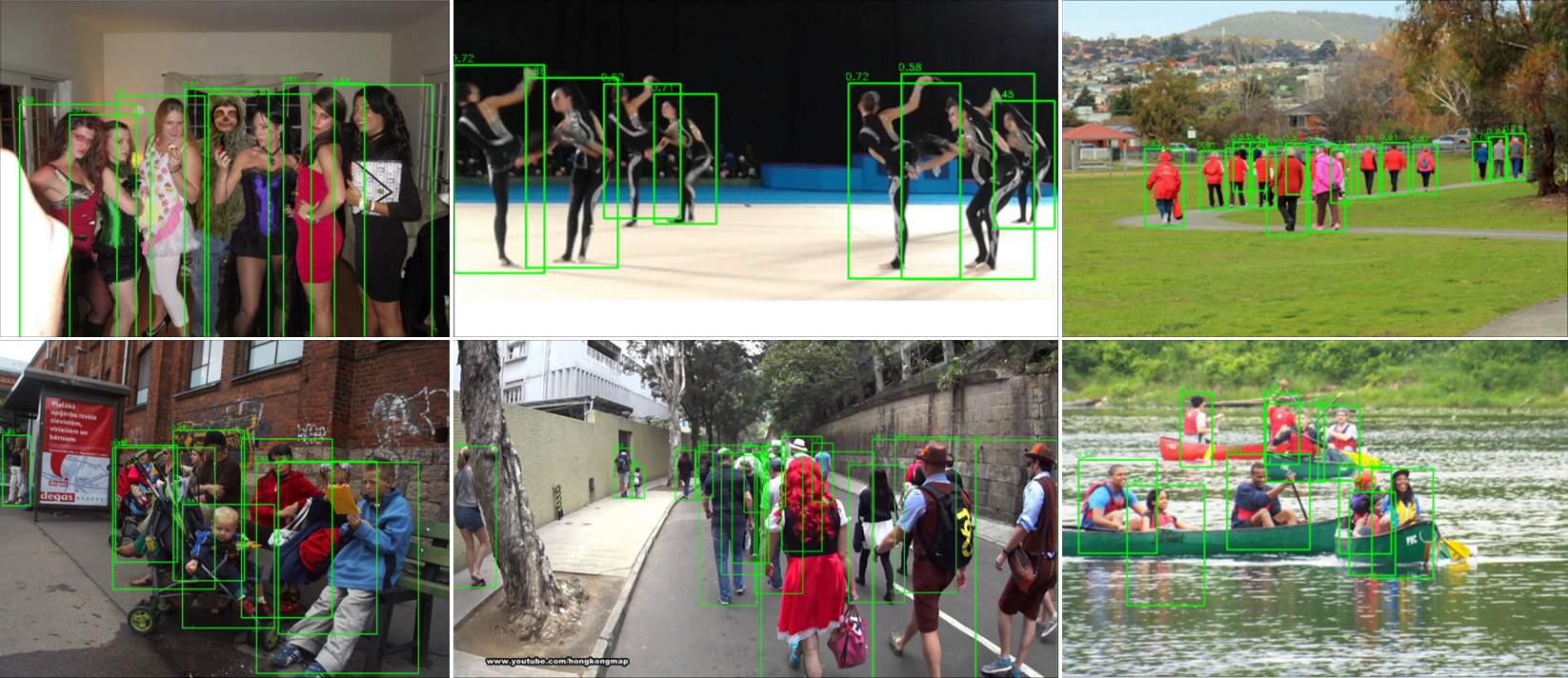}
\end{center}
\caption{Qualitative results on CrowdHuman human detection dataset, please zoom in to see some small detections.}
\label{figure-crowdhuman}
\end{figure*}

The results of the proposed SADet and other state-of-the-art methods on CrowdHuman dataset are reported in Table \ref{table8}. Based on the two-step regression structure, our SADet outperforms all the other one-stage detectors with $60.13\%$ MR$^{-2}$, $95.02\%$ Recall and $82.14\%$ AP on CrowdHuman, which not only demonstrates the superiority of our method, but also verifies its generalization capacity to other scenarios, \emph{i.e.}, human detection. Some qualitative results are depicted in Figure. \ref{figure-crowdhuman}, threshold is set to $0.4$.

\section{Conclusion}
In this paper, we propose a series of systematic optimization strategies for the detection pipeline of one-stage detector, forming a single shot anchor-based detector (SADet) for efficient and accurate pedestrian detection. Specifically, we first introduce a new design of soft label to the sample generation process to make full use of all valid samples and improve the robustness of classification. Then we optimize the design of $GIoU$ loss and further propose a new bounding box regression loss, named Center-$IoU$ loss, which not only alleviates the defects of $GIoU$ loss, but also enforces the predicted bounding boxes to be close to the associated objects and locate compactly, which is helpful to the precision improvement of bounding box regression. Meanwhile, for occluded pedestrian detection, we design Cosine-NMS for the postprocess of predictions to reduce false detections of the adjacent overlapping pedestrians, by assigning a higher penalty to the highly overlapped detections. Besides, the adaptive anchor matching is also proposed to enable the model to adaptively match the anchor boxes to full or visible bounding boxes according to the degree of occlusion, which further strengthens the ability of detecting occluded pedestrians. Our method is trained in an end-to-end fashion and achieves state-of-the-art accuracy on challenging pedestrian detection benchmarks, \emph{i.e.}, CityPersons\cite{zhang2017citypersons}, Caltech\cite{dollar2011pedestrian}, and human detection benchmark CrowdHuman\cite{shao2018crowdhuman} with real-time speed of $20$ FPS for VGA-resolution images.

\ifCLASSOPTIONcaptionsoff
  \newpage
\fi



\bibliographystyle{IEEEtran}

\end{document}